\documentclass[runningheads]{llncs}

\usepackage[T1]{fontenc}

\usepackage{graphicx,verbatim}

\usepackage[table]{xcolor}
\usepackage{multirow}
\usepackage{tabu}
\usepackage{bbm}
\usepackage{diagbox}
\usepackage[english]{babel}
\usepackage{comment}
\usepackage{array}
\usepackage{amstext}
\usepackage{makecell}
\usepackage{caption}
\usepackage{tablefootnote}
\usepackage{babel}
\usepackage{booktabs} 
\usepackage{color}
\usepackage{amsmath}
\usepackage{wrapfig}
\usepackage{amssymb}

\newif\ifdraft
\draftfalse
\drafttrue

\ifdraft
 \newcommand{\PF}[1]{{\color{red}{\bf PF: #1}}}
 
 \newcommand{\MS}[1]{{\color{green}{\bf MS: #1}}}
 
 \newcommand{\ZD}[1]{{\color{violet}{\bf ZD: #1}}}
 
 \newcommand{\YH}[1]{{\color{blue}{\bf YH: #1}}}
 
 \newcommand{\SPE}[1]{{\color{orange}{\bf SS: #1}}}
 
 \newcommand{\WJ}[1]{{\color{orange}{\bf WJ: #1}}}
 
  \newcommand{\JC}[1]{{\color{blue}{\bf JC: #1}}}
 
  \newcommand{\placeholder}[1]{{\color{green}{placeholder: #1}}}
\else
 \newcommand{\PF}[1]{}
 
 \newcommand{\KY}[1]{}
 
 \newcommand{\MS}[1]{}
 
 \newcommand{\ZD}[1]{}
 
 \newcommand{\YH}[1]{}
 
 \newcommand{\SPE}[1]{}
 
 \newcommand{\WJ}[1]{}
 
 \newcommand{\JC}[1]{}
 
 \newcommand{\placeholder}[1]{}
\fi

\newcommand{\parag}[1]{\vspace{-3mm}\paragraph{#1}}

\begin{document}

\title{DiffAtlas: GenAI-fying Atlas Segmentation via Image-Mask Diffusion}
\titlerunning{GenAI-fying Atlas Segmentation via Diffusion}

\authorrunning{H. Zhang et al.}

\author{
Hantao Zhang \inst{1,2}$^*$
\and
Yuhe Liu \inst{3}\thanks{Equal contribution. Conducted during the authors' research internships at EPFL.}
\and
Jiancheng Yang \inst{1}\thanks{{Corresponding author: J. Yang (\url{jiancheng.yang@epfl.ch}), who led the project. }}
\and
Weidong Guo \inst{2} 
\and \\
Xinyuan Wang \inst{3} 
\and
Pascal Fua\inst{1} 
}

\institute{Swiss Federal Institute of Technology Lausanne (EPFL), Lausanne, Switzerland
\\ \and
University of Science and Technology of China (USTC), Hefei, China
\\ \and
Beihang University, Beijing, China
}

\maketitle            

\begin{abstract}

Accurate medical image segmentation is crucial for precise anatomical delineation. Deep learning models like U-Net have shown great success but depend heavily on large datasets and struggle with domain shifts, complex structures, and limited training samples. Recent studies have explored diffusion models for segmentation by iteratively refining masks. However, these methods still retain the conventional image-to-mask mapping, making them highly sensitive to input data, which hampers stability and generalization. In contrast, we introduce DiffAtlas, a novel generative framework that models both images and masks through diffusion during training, effectively “GenAI-fying” atlas-based segmentation. During testing, the model is guided to generate a specific target image-mask pair, from which the corresponding mask is obtained. DiffAtlas retains the robustness of the atlas paradigm while overcoming its scalability and domain-specific limitations. Extensive experiments on CT and MRI across same-domain, cross-modality, varying-domain, and different data-scale settings using the MMWHS and TotalSegmentator datasets demonstrate that our approach outperforms existing methods, particularly in limited-data and zero-shot modality segmentation. Code is available at \url{https://github.com/M3DV/DiffAtlas}.


\end{abstract}
\section{Introduction}

In recent years, deep learning-based segmentation methods have achieved remarkable success in the field of medical image analysis~\cite{azad2024medical}. Standard segmentation architectures, such as the U-Net~\cite{ronneberger2015u}, are designed as feedforward neural networks and have demonstrated strong performance across a wide range of applications. The advantages of such feedforward models include their computational efficiency and ease of deployment. However, these methods also have notable limitations: they often struggle with fine-grained details, lack robustness to domain shifts, and suffer from complex morphological variations and continuity issues when faced with challenging scenarios. To address these issues, many approaches have been proposed, such as incorporating attention mechanisms~\cite{wang2022uctransnet}, multi-scale feature extraction~\cite{kushnure2021ms}, and adversarial training~\cite{nie2020adversarial}. Despite these efforts, challenges persist, especially with complex anatomical structures under limited samples~\cite{feng2021interactive} and multi-modal cross-domain issues~\cite{ding2022cross,guan2021domain}.

With the emergence of generative AI (GenAI)~\cite{kazerouni2023diffusion}, this novel paradigm has been introduced into many medical image analysis tasks~\cite{zhang2024lefusion}, such as segmentation~\cite{wu2024medsegdiff,chen2024hidiff} and anomaly detection~\cite{liu2025survey}, to enhance performance. For instance, Wu et al.~\cite{wu2024medsegdiff} incorporated image-based control into the diffusion to “GenAI-fy” segmentation. However, such diffusion-based methods~\cite{rahman2023ambiguous} are highly sensitive to input images and fail to maintain robustness when crossing domains. 


\begin{figure}[tb]
        \centering
	\includegraphics[width=1\linewidth]{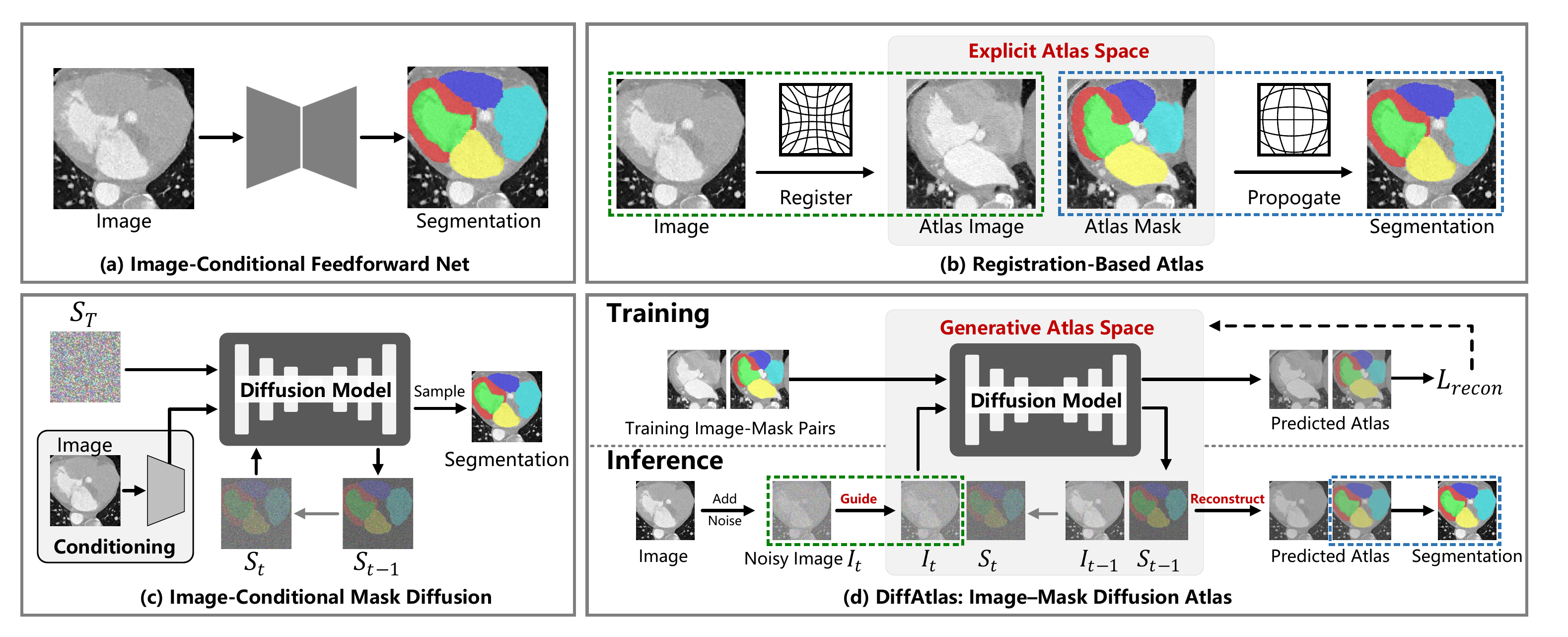}
    \caption{\textbf{Illustration of different segmentation paradigms.} (a) Directly maps the input image to the segmentation mask; (b) Uses registration to align a labeled atlas with the target image and propagate atlas labels; (c) Conditioned on the input image, achieves image-to-segmentation mask mapping using diffusion; (d) Train: Parameterizes an atlas and simultaneously generates image-mask pairs. Test: Uses the noisy image as guidance to generate the corresponding atlas.}
    \vspace{-10px}
	\label{fig:diffatlas_model}
\end{figure}

In this work, we revisit atlas-based segmentation~\cite{chen2010image,bai2015bi,ding2022cross}. This method, which involves image registration, label propagation, and refinement, was once the gold standard for medical image segmentation before the deep learning era~\cite{feuerstein2012mediastinal,zikic2014encoding}. It offers advantages such as interpretability, anatomical consistency, and robustness to small misalignments. However, atlas-based methods face key challenges: they are not easily scalable, struggle with fine-grained structures, and require labor-intensive, domain-specific atlases~\cite{zhuang2015multiatlas}. Despite these drawbacks, the robustness of atlas-based methods to misalignments remains a key advantage: even with misalignments, the segmentation remains meaningful~\cite{vakalopoulou2018atlasnet}. This stands in stark contrast to many feedforward neural network-based methods~\cite{chen2024transunet}, which often fail catastrophically when errors occur during segmentation.

In contrast, we propose a fundamentally different neural network segmentation approach based on the diffusion paradigm, parameterized for atlas-based segmentation, named \textbf{DiffAtlas}. Our method builds upon the principles of atlas-based registration and label propagation, while leveraging modern generative AI techniques to GenAI-fy the process. Specifically, DiffAtlas parameterizes the atlas as a generative model, capturing the joint distribution of \textbf{image-mask pairs} during training. During inference, DiffAtlas performs segmentation through a guided generative process, where the noisy image progressively replaces the predicted image at each step in the image-mask pair. This process \textbf{guides the pair toward the target}, generating the corresponding mask. We conduct extensive experiments on two whole-heart segmentation datasets: MM-WHS~\cite{zhuang2019evaluation} and TotalSegmentator~\cite{zhuang2019evaluation}, covering various settings including base segmentation, few-shot and extremely few-shot sample settings, as well as zero-shot cross-modality segmentation. Our results demonstrate that the proposed method achieves state-of-the-art performance in these scenarios.

\section{Method}

Atlas-based segmentation methods are well-known for their ability to enforce anatomical consistency and preserve global structural integrity in segmentation masks\cite{ranem2024continual}. However, their reliance on predefined atlases and computationally intensive registration steps makes them inherently inflexible and difficult to scale to diverse anatomical variations\cite{gibbons2023clinical}.

To tackle these challenges, we propose DiffAtlas, a novel framework that combines atlas priors with the flexibility of generative diffusion models, as illustrated in Fig.~\ref{fig:diffatlas_model}(d). DiffAtlas constructs a generative atlas space through a learned diffusion process, which implicitly encodes diverse anatomical structures. This eliminates the need for explicit atlas registration while preserving the anatomical consistency central to atlas-based methods.

By modeling the joint prior of images and segmentation masks, DiffAtlas generates anatomically consistent and tightly aligned pairs. During inference, a noisy image (pre-noised from the input image at the corresponding time step $t$) replaces the predicted image, guiding the model to output the target pair and, consequently, the corresponding mask. Through this, DiffAtlas seamlessly integrates the target image’s global anatomical priors into the inference process on any dataset, enabling segmentation without the need for additional training. At the same time, its reliance on a learned generative atlas space, rather than explicit registration, significantly reduces both time and spatial complexity, ensuring greater flexibility and scalability, and making it a robust solution.

\subsection{Revisiting Previous Methods}

\paragraph{Image-Conditional Feedforward Network.}

As illustrated in Fig.~\ref{fig:diffatlas_model}(a), feedforward neural networks, such as U-Net~\cite{ronneberger2015u}, are widely used for image segmentation tasks.These models learn a deterministic mapping from the input image $\mathbb{I}$ to the segmentation mask $\mathbb{S}$, denoted as $f_\theta: \mathbb{I} \rightarrow \mathbb{S}$. The training objective typically minimizes the cross-entropy loss, dice loss or their combination. While efficient, these methods entirely rely on the input image, making them highly sensitive to input conditions. For instance, noise, artifacts, or incomplete information in the input image can significantly degrade segmentation accuracy. Furthermore, the lack of global anatomical priors limits the ability of these methods to enforce structural consistency, leading to poor robustness and generalization, particularly for complex anatomical regions or across diverse datasets~\cite{ma2024segment}.

\parag{Image-Conditional Mask Diffusion.}

Image-Conditional Mask Diffusion extends the image-conditional paradigm by modeling the conditional distribution $p(\mathbf{S}|\mathbf{I})$, where $\mathbf{S}$ is the segmentation mask and $\mathbf{I}$ is the input image. By introducing a probabilistic framework, Image-Conditional Mask Diffusion decomposes the generation of the segmentation mask into a step-by-step denoising process. The forward process $q(\mathbf{S}_t|\mathbf{S}_{t-1})$ progressively corrupts the ground-truth mask $\mathbf{S}$ by adding Gaussian noise. Specifically, the noisy mask $\mathbf{S}_t$ at step $t$ is obtained as:

\begin{equation}
\mathbf{S}_t = \sqrt{\alpha_t} \mathbf{S}_{t-1} + \sqrt{1-\alpha_t} \boldsymbol{\epsilon}, \quad \boldsymbol{\epsilon} \sim \mathcal{N}(\mathbf{0}, \mathcal{I}),
\end{equation}

where $\alpha_t$ is a predefined noise schedule that controls the balance between the signal and noise, and $\mathcal{I}$ is the identity matrix.

The reverse process $p_\theta(\mathbf{S}_{t-1}|\mathbf{S}_t, \mathbf{I})$ iteratively refines the noisy mask by learning the conditional distribution:

\begin{equation}
p_\theta(\mathbf{S}_{t-1}|\mathbf{S}_t, \mathbf{I}) = \mathcal{N}(\mathbf{S}_{t-1}; \mathbf{\mu}_\theta(\mathbf{S}_t, \mathbf{I}, t), \mathbf{\Sigma}_\theta(\mathbf{S}_t, \mathbf{I}, t)),
\end{equation}

where $\mathbf{\mu}_\theta$ and $\mathbf{\Sigma}_\theta$ are the learned mean and covariance functions. The introduction of diffusion enables this paradigm to generate plausible masks through a step-by-step process. However, it remains fundamentally tied to the image-conditional paradigm, inheriting its limitations in the mapping from image to mask.

\parag{Registration-Based Atlas.}

Atlas-based methods address the lack of global priors by leveraging predefined labeled atlases $(\mathbf{A}_I, \mathbf{A}_S)$. These methods register the atlas image $\mathbf{A}_I$ to the target image $\mathbf{I}$, propagating the atlas labels $\mathbf{A}_S$ through the deformation field $\mathbf{\phi}^*$:
\begin{equation}
\mathbf{S} = \mathbf{A}_S \circ \mathbf{\phi}^*,
\end{equation}
where the deformation field $\mathbf{\phi}^*$ is computed by minimizing an energy function:
\begin{equation}
\mathbf{\phi}^* = \arg\min_{\mathbf{\phi}} \mathcal{D}(\mathbf{I}, \mathbf{A}_I \circ \mathbf{\phi}) + \lambda \mathcal{R}(\mathbf{\phi}),
\end{equation}
with $\mathcal{D}$ measuring similarity between $\mathbf{I}$ and the deformed atlas $\mathbf{A}_I \circ \mathbf{\phi}$, and $\mathcal{R}$ being a regularization term. Although effective in enforcing anatomical consistency, these methods are computationally expensive and heavily depend on the quality of the predefined atlas. Their performance can degrade significantly when the atlas fails to represent diverse anatomical variations.

\subsection{DiffAtlas: Image–Mask Diffusion Atlas}

\paragraph{Joint Image-Mask Prior as Generative Atlas Space.}

DiffAtlas models the input image and segmentation mask as a pair $(\mathbf{I}, \mathbf{S})$ within a generative atlas space, learned through a diffusion process. Unlike traditional methods that rely on predefined atlases and explicit registration, this learned space captures global anatomical consistency while remaining flexible to diverse anatomical variations. By training the model to reconstruct clean pairs $(\mathbf{I}, \mathbf{S})$ from noisy versions $(\mathbf{I}_t, \mathbf{S}_t)$, DiffAtlas learns the joint distribution $p_\theta(\mathbf{I}, \mathbf{S})$. This ensures that the segmentation mask $\mathbf{S}$ is both anatomically consistent and naturally aligned with its corresponding image $\mathbf{I}$.

\parag{Noisy Image Guidance for Input-Conditioned Sampling.}

To ensure that the generated image-mask pair aligns with the input image during inference, DiffAtlas employs a noisy image replacement strategy. During the replacement process, the noisy version of the input image $\mathbf{I}_\text{input}$ is introduced at each timestep $t$ to guide the sampling process. Specifically, the noisy image $\mathbf{I}_t$ is replaced with:
\begin{equation}
\mathbf{I}_t = \sqrt{\bar{\alpha}_t} \mathbf{I}_\text{input} + \sqrt{1-\bar{\alpha}_t} \boldsymbol{\epsilon},
\end{equation}
where $\boldsymbol{\epsilon}$ is Gaussian noise sampled from $\mathcal{N}(\mathbf{0}, \mathcal{I})$, and $\bar{\alpha}_t$ is the cumulative product of the noise schedule $\alpha_t$ up to timestep $t$. By conditioning the diffusion process on the input image, this guidance ensures that the sampled segmentation mask $\mathbf{S}$ aligns closely with the anatomical structures of $\mathbf{I}_\text{input}$, while preserving the global consistency enforced by the generative atlas space.

\subsection{DiffAtlas Implementation}
\paragraph{Training Process.}

During training, DiffAtlas optimizes a neural network to predict the added noise $\boldsymbol{\epsilon}$ on both the image $\mathbf{I}$ and the segmentation mask $\mathbf{S}$ in the forward diffusion process. The objective minimizes the mean squared error between the predicted noise $\boldsymbol{\epsilon}_\theta$ and the true noise $\boldsymbol{\epsilon}$:
\begin{equation}
\mathcal{L}_{\text{train}} = \mathbb{E}_{(\mathbf{I}, \mathbf{S}), t, \boldsymbol{\epsilon}} \left[\|\boldsymbol{\epsilon} - \boldsymbol{\epsilon}_\theta(\mathbf{I}_t, \mathbf{S}_t, t)\|^2\right].
\end{equation}

Modeling joint distributions of image-mask pairs has been explored in prior works, such as for representation learning or feature extraction~\cite{sauvalle2024hybrid}. We leverage this to construct a generative atlas space. Our primary focus is on guiding matching within the implicit atlas space for image segmentation. Furthermore, to enhance spatial continuity and structural coherence, DiffAtlas represents the mask $\mathbf{S}$ using a signed distance function (SDF), which encodes the distance to the mask boundary~\cite{bogensperger2024flowsdf}, improving the capture of fine anatomical details and ensuring smooth transitions between regions.

\parag{Inference Process.}

During inference, DiffAtlas starts with a randomly initialized noisy image-mask pair $(\mathbf{I}_T, \mathbf{S}_T)$ and refines it iteratively through the reverse diffusion process over $T$ timesteps. At each timestep $t$, the noisy image $\mathbf{I}_t$ is replaced with the noisy version of the input image $\mathbf{I}_\text{input}$. This replacement gently anchors the generative process to the anatomical structures of the input, ensuring that the model remains guided by its features throughout the denoising process.

As the diffusion unfolds, the noisy pair $(\mathbf{I}_t, \mathbf{S}_t)$ progressively transforms into a clean and coherent pair $(\mathbf{I}, \mathbf{S})$. In this final pair, the generated image $\mathbf{I}$ naturally converges to closely resemble the input image $\mathbf{I}_\text{input}$, while the segmentation mask $\mathbf{S}$ emerges as an inherently compatible counterpart. Without any explicit enforcement, the mask $\mathbf{S}$ aligns seamlessly with the anatomical structures of $\mathbf{I}_\text{input}$, reflecting both global consistency and local precision. This effortless alignment arises naturally from the learned joint distribution in the generative atlas space, where the anatomical priors ensure harmony between the image and mask. By combining the implicit guidance of the generative prior with direct conditioning on the input image, DiffAtlas achieves segmentation results that are both precise and intuitively consistent.

\section{Experiments}
\subsection{Setup}

\begin{table}[tb]
	\caption{\textbf{Segmentation under full training setting.}}
	\label{tab:seg}
	\centering
        {\scriptsize

    \begin{tabular*}{\hsize}{@{}@{\extracolsep{\fill}}lcccccccccccc@{}}
        
        \toprule
        \multirow{2}{*}{Methods} & \multicolumn{2}{c}{Myo} & \multicolumn{2}{c}{LV} & \multicolumn{2}{c}{LA} & \multicolumn{2}{c}{RA} & \multicolumn{2}{c}{RV} & \multicolumn{2}{c}{Average} \\
        & Dice & NSD & Dice & NSD & Dice & NSD & Dice & NSD & Dice & NSD & Dice & NSD \\
        \midrule
        \multicolumn{12}{l}{\textit{\textbf{Full} TS training set $
        \rightarrow$ TS test set} } \\
        (a) ICF~\cite{isensee2021nnu} & 80.22 & 83.67 & \cellcolor{red!25}78.06 & 75.16 & 88.32 & 82.24 & 83.21 & 73.43 & \cellcolor{red!25}88.48 & 81.62 & 83.66 & 79.22 \\
        (b) RBA~\cite{ding2022cross} & NaN & NaN & NaN & NaN & NaN & NaN & NaN & NaN & NaN & NaN & NaN & NaN \\
        (c) ICMD~\cite{wu2024medsegdiff} & 77.80 & 84.30 & 74.20 & 73.34 & 87.44 & 81.64 & 82.72 & 76.25 & 86.53 & 79.68 & 81.74 & 79.04 \\
        (d) DA (Ours) & \cellcolor{red!25}83.52 & \cellcolor{red!25}88.87 & 77.21 & \cellcolor{red!25}75.70 & \cellcolor{red!25}90.13 & \cellcolor{red!25}86.35 & \cellcolor{red!25}86.52 & \cellcolor{red!25}80.76 & 88.36 & \cellcolor{red!25}82.06 & \cellcolor{red!25}85.17 & \cellcolor{red!25}82.74 \\
        \midrule
        \multicolumn{12}{l}{\textit{\textbf{Full} MMWHS-CT training set $
        \rightarrow$ MMWHS-CT test set}} \\
        (a) ICF~\cite{isensee2021nnu} & 57.88 & 47.18 & 60.05 & 22.41 & 73.31 & 42.62 & 49.68 & 25.35 & 69.24 & 43.25 & 62.03 & 36.16 \\
        (b) RBA~\cite{ding2022cross} & 51.13 & 40.56 & 65.28 & 30.66 & 60.04 & 31.13 & 66.96 & 39.54 & 55.38 & 29.47 & 59.76 & 34.27 \\
        (c) ICMD~\cite{wu2024medsegdiff} & 58.03 & 52.63 & 41.98 & 26.05 & 68.51 & 45.64 & 66.47 & 41.81 & 39.98 & 25.56 & 54.99 & 38.34 \\
        (d) DA (Ours) & \cellcolor{red!25}78.02 & \cellcolor{red!25}73.75 & \cellcolor{red!25}87.27 & \cellcolor{red!25}74.74 & \cellcolor{red!25}86.68 & \cellcolor{red!25}74.41 & \cellcolor{red!25}80.25 & \cellcolor{red!25}65.87 & \cellcolor{red!25}84.02 & \cellcolor{red!25}66.79 & \cellcolor{red!25}83.25 & \cellcolor{red!25}71.11 \\

        \midrule
        \multicolumn{12}{l}{\textit{\textbf{Full} MMWHS-MRI training set $
        \rightarrow$ MMWHS-MRI test set}} \\
        (a) ICF~\cite{isensee2021nnu} & 41.11 & 38.90 & 51.63 & 20.28 & 67.63 & 31.98 & 49.66 & 22.00 & 50.05 & 28.99 & 52.02 & 28.43 \\
        (b) RBA~\cite{ding2022cross} & 38.28 & 42.01 & 53.10 & 30.49 & 63.19 & 34.32 & 59.34 & 28.18 & 49.76 & 26.84 & 52.73 & 32.37 \\
        (c) ICMD~\cite{wu2024medsegdiff} & 53.14 & \cellcolor{red!25}55.40 & 32.97 & 21.98 & 65.51 & 40.97 & 61.84 & 37.48 & 39.87 & 27.10 & 50.67 & 36.59 \\
        (d) DA (Ours) & \cellcolor{red!25}57.97 & 46.42 & \cellcolor{red!25}70.67 & \cellcolor{red!25}41.11 & \cellcolor{red!25}75.70 & \cellcolor{red!25}43.93 & \cellcolor{red!25}73.11 & \cellcolor{red!25}44.14 & \cellcolor{red!25}65.77 & \cellcolor{red!25}41.64 & \cellcolor{red!25}68.64 & \cellcolor{red!25}43.45 \\
        \bottomrule
    \end{tabular*}
    }
\end{table}

\paragraph{Dataset.} 
This work uses two datasets: MM-WHS~\cite{zhuang2019evaluation} and TotalSegmentator (TS)~\cite{zhuang2019evaluation} dataset. For both datasets, we retained five labels for full heart segmentation: Myocardium of the LV (Myo), LV blood cavity (LV), left atrium (LA), right atrium (RA), and right ventricle (RV). We used the cardiac CT portion of the TS, which contains 746 cases, reserving 20\% for testing. For MM-WHS, we used 20 available CT cases and 20 MRI cases. 

The specific experimental settings are as follows: \textbf{Full setting:} For the TS, MMWHS-CT, and MMWHS-MRI datasets, we used 80\% for training and 20\% for testing. \textbf{Few-shot setting:} The test set remains the same as in the previous setting, comprising 20\% of the total, with two configurations: 2-shot (2 training samples) and 4-shot (4 training samples). \textbf{Zero-shot cross-modality setting:} We used the same train-test split, dividing the settings into three groups: large data with different source CT to MRI, small data with same-source CT to MRI, and MRI to CT, as shown in Tab.~\ref{tab:cross_seg}.

\parag{Method Comparison.} We compared our method with publicly available state-of-the-art methods from each category shown in Fig.\ref{fig:diffatlas_model}, including \textbf{(a) ICF} (Image-Conditional Feedforward): nnU-Net\cite{isensee2021nnu}; \textbf{(b) RBA} (Registration-Based Atlas): CMMAS~\cite{ding2022cross}; \textbf{(c) ICMD} (Image-Conditional Mask Diffusion): MedSegDiffv2~\cite{wu2024medsegdiff}; \textbf{(d) DA (Ours)}: DiffAtlas. For fairness, all models use the same data augmentation, with no additional augmentation for nnU-Net.

\subsection{Performance Evaluation}
For all experiments, we followed Metrics Reloaded~\cite{maier2024metrics} and related works~\cite{ding2022cross,wu2024medsegdiff} and employed the Dice similarity coefficient (Dice) and normalized surface distance (NSD)~\cite{deepmind2018surface} as metrics. We use \colorbox{red!25}{red} to highlight the best.

\paragraph{Full Training.} As shown in Tab.~\ref{tab:seg}, under the same-dataset setting, our proposed (d) DA (DiffAtlas) outperforms state-of-the-art methods, including (a) ICF (nnU-Net\cite{isensee2021nnu}), on both the large-scale CT TS dataset and the smaller MMWHS CT and MRI datasets, demonstrating its effectiveness and scalability. However, (b) RBA (Registration-Based Atlas): CMMAS~\cite{ding2022cross} is limited by the high time and space complexity of the atlas paradigm, restricting its scalability. As a result, it could not be executed on the large-scale TS dataset, with results marked as ‘NaN’.

\parag{Few-shot Training.} 
\begin{table}[tb]
	\caption{\textbf{Segmentation under few-shot training setting.}}
	\label{tab:few_seg}
	\centering
    {\scriptsize 
    \begin{tabular*}{\hsize}{@{}@{\extracolsep{\fill}}lcccccccccccc@{}}
        \toprule
        \multirow{2}{*}{Methods} & \multicolumn{2}{c}{Myo} & \multicolumn{2}{c}{LV} & \multicolumn{2}{c}{LA} & \multicolumn{2}{c}{RA} & \multicolumn{2}{c}{RV} & \multicolumn{2}{c}{Average} \\
        & Dice & NSD & Dice & NSD & Dice & NSD & Dice & NSD & Dice & NSD & Dice & NSD \\
        \midrule    
        \multicolumn{12}{l}{\textit{\textbf{2-shot} MMWHS-CT training samples $
        \rightarrow$ MMWHS-CT test set}} \\
        (a) ICF~\cite{isensee2021nnu} & 26.09 & 33.54 & 49.66 & 16.93 & 69.70 & 42.22 & 39.16 & 21.94 & 50.17 & 15.83 & 46.96 & 26.09 \\
        (b) RBA~\cite{ding2022cross} & 44.29 & 39.16 & 58.83 & 13.85 & 67.99 & 31.68 & 61.83 & 33.45 & 59.83 & 32.70 & 58.55 & 30.17 \\
        (c) ICMD~\cite{wu2024medsegdiff} & 38.26 & 39.59 & 38.31 & 24.22 & 49.05 & 33.38 & 50.91 & 25.70 & 29.31 & 19.08 & 41.17 & 28.39 \\
        (d) DA (Ours) & \cellcolor{red!25}70.30 & \cellcolor{red!25}58.45 & \cellcolor{red!25}80.86 & \cellcolor{red!25}59.60 & \cellcolor{red!25}82.39 & \cellcolor{red!25}45.30 & \cellcolor{red!25}78.01 & \cellcolor{red!25}58.42 & \cellcolor{red!25}77.08 & \cellcolor{red!25}55.88 & \cellcolor{red!25}77.73 & \cellcolor{red!25}55.53 \\
        \midrule
        \multicolumn{12}{l}{\textit{\textbf{4-shot} MMWHS-CT training samples $
        \rightarrow$ MMWHS-CT test set}} \\
        (a) ICF~\cite{isensee2021nnu} & 35.35 & 31.13 & 60.25 & 18.71 & 71.78 & 31.14 & 40.40 & 16.29 & 64.39 & 40.73 & 54.43 & 27.60 \\
        (b) RBA~\cite{ding2022cross} & 49.61 & 41.69 & 61.03 & 34.53 & 71.21 & 35.72 & 58.01 & 39.12 & 65.95 & 33.88 & 61.16 & 36.99 \\
        (c) ICMD~\cite{wu2024medsegdiff} & 43.32 & 40.74 & 38.49 & 24.55 & 62.80 & 38.74 & 51.05 & 25.59 & 28.65 & 19.34 & 44.86 & 29.79 \\
        (d) DA (Ours) & \cellcolor{red!25}72.05 & \cellcolor{red!25}53.21 & \cellcolor{red!25}83.07 & \cellcolor{red!25}56.69 & \cellcolor{red!25}86.72 & \cellcolor{red!25}44.40 & \cellcolor{red!25}74.56 & \cellcolor{red!25}41.35 & \cellcolor{red!25}79.50 & \cellcolor{red!25}55.69 & \cellcolor{red!25}79.18 & \cellcolor{red!25}50.27 \\
        
        \midrule
        \multicolumn{12}{l}{\textit{\textbf{2-shot} MMWHS-MRI training samples $
        \rightarrow$ MMWHS-MRI test set}} \\
        (a) ICF~\cite{isensee2021nnu} & 25.95 & 27.15 & 23.63 & 13.68 & 35.33 & 16.32 & 37.52 & 18.47 & 13.16 & 12.73 & 27.12 & 17.67 \\
        (b) RBA~\cite{ding2022cross} & 33.74 & 35.17 & 40.24 & \cellcolor{red!25}25.99 & 51.59 & 28.73 & 61.10 & 33.63 & 38.38 & 23.38 & 45.01 & 29.38 \\
        (c) ICMD~\cite{wu2024medsegdiff} & 45.39 & \cellcolor{red!25}44.72 & 7.90 & 6.97 & \cellcolor{red!25}61.35 & \cellcolor{red!25}36.69 & 56.13 & \cellcolor{red!25}33.71 & 22.43 & 17.91 & 38.64 & 28.00 \\
        (d) DA (Ours) & \cellcolor{red!25}49.56 & 42.06 & \cellcolor{red!25}57.48 & 24.71 & 61.05 & 28.52 & \cellcolor{red!25}64.59 & 30.19 & \cellcolor{red!25}51.26 & \cellcolor{red!25}34.55 & \cellcolor{red!25}56.79 & \cellcolor{red!25}32.01 \\
        
        \midrule
        
        \multicolumn{12}{l}{\textit{\textbf{4-shot} MMWHS-MRI training samples $
        \rightarrow$ MMWHS-MRI test set}} \\
        (a) ICF~\cite{isensee2021nnu} & 33.12 & 32.30 & 48.79 & 25.47 & 55.48 & 26.19 & 41.42 & 19.53 & 31.78 & 19.57 & 42.12 & 24.61 \\
        (b) RBA~\cite{ding2022cross} & 47.88 & 43.62 & 49.24 & 24.56 & 59.99 & 30.42 & 65.89 & 35.18 & 49.78 & 28.41 & 54.56 & 32.44 \\
        (c) ICMD~\cite{wu2024medsegdiff} & 49.19 & 44.88 & 20.64 & 16.48 & 65.81 & 35.22 & 55.19 & 33.74 & 9.44 & 17.09 & 40.06 & 29.48 \\
        (d) DA (Ours) & \cellcolor{red!25}54.34 & \cellcolor{red!25}48.42 & \cellcolor{red!25}69.69 & \cellcolor{red!25}39.43 & \cellcolor{red!25}72.09 & \cellcolor{red!25}44.78 & \cellcolor{red!25}75.92 & \cellcolor{red!25}48.09 & \cellcolor{red!25}63.22 & \cellcolor{red!25}41.46 & \cellcolor{red!25}67.05 & \cellcolor{red!25}44.44 \\
        \bottomrule
    \end{tabular*}
    }
\end{table}

As illustrated in Tab.~\ref{tab:few_seg}, atlas-based methods ((b) and (d)) demonstrate greater robustness in small-data learning than image-mask mapping methods ((a) and (c)). DiffAtlas (d) achieves the best overall performance, followed by (b), leveraging the atlas-based registration paradigm. By parameterizing the atlas and modeling both image and mask, it utilizes diffusion’s ability to enhance learning with limited samples.

\parag{Zero-shot Cross-domain Evaluation.} 

\begin{table}[tb]
	\caption{\textbf{Segmentation under zero-shot cross-domain evaluation setting.}}
	\label{tab:cross_seg}
	\centering
    {\scriptsize 
    \begin{tabular*}{\hsize}{@{}@{\extracolsep{\fill}}lcccccccccccc@{}}
        \toprule
        \multirow{2}{*}{Methods} & \multicolumn{2}{c}{Myo} & \multicolumn{2}{c}{LV} & \multicolumn{2}{c}{LA} & \multicolumn{2}{c}{RA} & \multicolumn{2}{c}{RV} & \multicolumn{2}{c}{Average} \\
        & Dice & NSD & Dice & NSD & Dice & NSD & Dice & NSD & Dice & NSD & Dice & NSD \\
        \midrule
        \multicolumn{12}{l}{\textit{Full TS training set ONLY $
        \rightarrow$ MMWHS-MRI dataset} } \\
        (a) ICF~\cite{isensee2021nnu} & 28.36 & 34.71 & 45.57 & 21.13 & 51.66 & 22.89 & 50.75 & 24.71 & 45.25 & 22.80 & 44.32 & 25.25 \\
        (b) RBA~\cite{ding2022cross} & 24.13 & 29.63 & 40.79 & 24.28 & 46.09 & 23.59 & 39.48 & 22.98 & 42.03 & 24.02 & 38.51 & 24.90 \\
        (c) ICMD~\cite{wu2024medsegdiff}  & 29.26 & 24.01 & 75.53 & 62.06 & 57.86 & 30.02 & 66.18 & 43.26 & 47.74 & 37.07 & 55.31 & 39.29 \\
        (d) DA (Ours) & \cellcolor{red!25}62.97 & \cellcolor{red!25}52.31 & \cellcolor{red!25}81.49 & \cellcolor{red!25}66.65 & \cellcolor{red!25}83.25 & \cellcolor{red!25}37.85 & \cellcolor{red!25}80.13 & \cellcolor{red!25}57.95 & \cellcolor{red!25}74.37 & \cellcolor{red!25}53.89 & \cellcolor{red!25}76.44 & \cellcolor{red!25}53.73 \\
        \midrule
        \multicolumn{12}{l}{\textit{Full MMWHS-CT dataset ONLY $
        \rightarrow$ MMWHS-MRI dataset}} \\
        (a) ICF~\cite{isensee2021nnu} & 26.89 & 28.38 & 45.95 & 19.39 & 41.96 & 19.34 & 31.33 & 13.69 & 30.18 & 16.34 & 35.26 & 19.42 \\
        (b) RBA~\cite{ding2022cross} & 31.70 & \cellcolor{red!25}34.53 & 42.01 & 24.88 & 50.71 & 27.71 & 43.05 & 29.92 & 39.72 & \cellcolor{red!25}23.55 & 41.44 & 28.12 \\
        (c) ICMD~\cite{wu2024medsegdiff}  & 14.09 & 14.56 & 48.60 & \cellcolor{red!25}37.04 & 41.35 & 22.60 & 38.93 & 27.47 & 16.55 & 16.82 & 31.90 & 23.70 \\
        (d) DA (Ours) & \cellcolor{red!25}46.43 & 24.80 & \cellcolor{red!25}61.59 & 30.55 & \cellcolor{red!25}70.42 & \cellcolor{red!25}38.13 & \cellcolor{red!25}65.88 & \cellcolor{red!25}33.14 & \cellcolor{red!25}51.79 & 21.08 & \cellcolor{red!25}59.22 & \cellcolor{red!25}29.54 \\
        
        \midrule
        \multicolumn{12}{l}{\textit{Full MMWHS-MRI dataset ONLY $
        \rightarrow$ MMWHS-CT dataset}} \\
        (a) ICF~\cite{isensee2021nnu} & 25.03 & 30.29 & 47.54 & 20.01 & 56.74 & \cellcolor{red!25}24.12 & 52.59 & 21.85 & 48.46 & 18.22 & 46.07 & 22.90 \\
        (b) RBA~\cite{ding2022cross} & 32.44 & 32.61 & 51.43 & 23.30 & 52.76 & 23.18 & 51.26 & 26.25 & 40.43 & 23.06 & 45.67 & 25.68 \\
        (c) ICMD~\cite{wu2024medsegdiff} & 46.52 & 20.84 & 41.53 & 11.84 & \cellcolor{red!25}67.41 & 18.54 & 66.62 & 18.21 & 16.25 & 4.91 & 47.67 & 14.87 \\
        (d) DA (Ours)  & \cellcolor{red!25}54.24 & \cellcolor{red!25}33.72 & \cellcolor{red!25}72.22 & \cellcolor{red!25}36.97 & 61.90 & 21.58 & \cellcolor{red!25}69.14 & \cellcolor{red!25}32.06 & \cellcolor{red!25}61.49 & \cellcolor{red!25}27.78 & \cellcolor{red!25}63.80 & \cellcolor{red!25}30.42 \\
        
        \bottomrule
    \end{tabular*}
    }
\end{table}

Tab.~\ref{tab:few_seg} presents results across three settings with varying pretraining sample sizes, domains, and sources. Similar to the small-sample experiments, atlas-based methods ((b) and (d)) exhibit superior performance and robustness. DiffAtlas consistently achieves the best results, largely due to its guided inference process, which enhances target image matching and mitigates semantic gaps from domain shifts.

\subsection{Visual Analysis}
\begin{figure}[tb]
        \centering
	\includegraphics[width=1\linewidth]{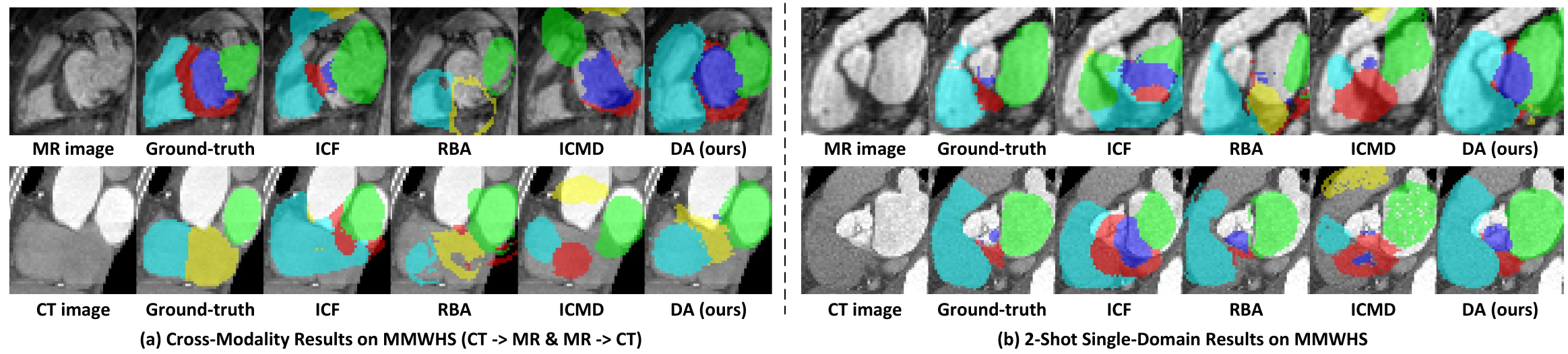}
    \caption{\textbf{Visualization of 2-shot cross-domain and few-shot results.}}
    \vspace{-10px}
	\label{fig:vis}
\end{figure}

Fig.~\ref{fig:vis} displays the segmentation results with Myo (red), LV (green), LA (dark blue), RA (yellow), and RV (light blue). In cross-domain scenarios, the first row of Fig.\ref{fig:vis}(a) demonstrates zero-shot segmentation from CT pretraining to MRI, while the second row shows MRI to CT. The results indicate that atlas-based methods (RBA and DA (ours)) maintain basic anatomical integrity across domains, avoiding significant misclassifications, such as the mis-segmentation of Myo (red) and RA (yellow) shown in ICMD. In 2-shot scenario, our method ensures the most accurate segmentation by preserving contour integrity.

\section{Conclusion}
In conclusion, we present DiffAtlas, which jointly models image-mask pairs during training and guides the model to generate target image-mask pairs, ultimately producing the corresponding masks during testing. This approach alleviates the limitations of the atlas paradigm, such as scalability challenges and struggles with fine-grained structures and domain-specific atlases, while preserving its advantages of anatomical consistency and robustness to small misalignments. We thoroughly explore its effectiveness across same-domain settings, different modalities, varying domains, and different data scales.

\bibliographystyle{splncs04}
\bibliography{string,reference}

 \end{document}